\documentclass[conference]{IEEEtran}
\IEEEoverridecommandlockouts

\DeclareUnicodeCharacter{0301}{\'{e}}

\usepackage{cite}
\usepackage{amsmath,amssymb,amsfonts}
\usepackage{algorithmic}
\usepackage{graphicx}
\usepackage{textcomp}
\usepackage{xcolor}
\def\BibTeX{{\rm B\kern-.05em{\sc i\kern-.025em b}\kern-.08em
    T\kern-.1667em\lower.7ex\hbox{E}\kern-.125emX}}
\begin{document}
\title{Yield Evaluation of Citrus Fruits based on the YoloV5 compressed by Knowledge Distillation  \\
{\footnotesize \textsuperscript{}}
\author{
\IEEEauthorblockN{
Yuqi Li\IEEEauthorrefmark{2},
Yuting He\IEEEauthorrefmark{2},
Yihang Zhou,Zirui Gong and
Renjie Huang}
\IEEEauthorblockA{College of Computer and Information Science College of Software, Southwest University, Chongqing, China}
\IEEEauthorblockA{lllyq13@email.swu.edu.cn, heyuting188@163.com, swuyihangzhou@163.com,gongzir5@gmail.com}
\IEEEauthorblockA{Corresponding Author: Renjie Huang \quad Email: Huangrj@swu.edu.cn}
\IEEEauthorblockA{\IEEEauthorrefmark{3}Yuqi Li and Yuting He are co-first authors.}}
}

\maketitle

\begin{abstract}
In the field of planting fruit trees, pre-harvest estimation of fruit yield is important for fruit storage and price evaluation. However, considering the cost, the yield of each tree cannot be assessed by directly picking the immature fruit. Therefore, the problem is a very difficult task. In this paper, a fruit counting and yield assessment method based on computer vision is proposed for citrus fruit trees as an example. Firstly, images of single fruit trees from different angles are acquired and the number of fruits is detected using a deep Convolutional Neural Network model YOLOv5, and the model is compressed using a knowledge distillation method. Then, a linear regression method is used to model yield-related features and evaluate yield. Experiments show that the proposed method can accurately count fruits and approximate the yield.
\end{abstract}

\begin{IEEEkeywords}
Yolov5, knowledge distillation, citrus fruits, object detection, linear regression
\end{IEEEkeywords}

\section{Introduction}
At present, most of the citrus fruits plantations in China and abroad use the obsolete method of manual yield measurement, which is not only labor-intensive, but also difficult to guarantee its accuracy. This traditional method of yield estimation also tests the experience of the estimator, and it is difficult to apply the experience of yield estimation from this specific fruit to other crops. Therefore, yield estimation is a very difficult problem~\cite{zekri2011factors}.

More research on fruit detection and yield measurement has also been carried out at home and abroad. The previous research steps for target detection mainly include methods such as image acquisition, color space conversion, model fusion, cutting for citrus fruits background, feature extraction, fitting, and classification~\cite{rodriguez2017machine,chen2022citrusyolo,ye2007prediction,khan2020agricultural}. These methods are relatively too complicated, some need to take photos in multiple views to make predictions, some need complex operations to apply to other varieties of fruit detection, and most of them are not ideal for fruit recognition under the influence of overlap, occlusion, small targets, etc. There are also problems such as less public datasets and difficulty in obtaining image data. 

To solve the above problems we propose a network for counting and yield prediction of citrus fruit trees based on image analysis~\cite{chen2022citrusyolo}. By collecting a large number of images of ripe citrus fruit and inputting them into the established deep learning model for training, thus citrus fruit is subjected to feature extraction to achieve the function of citrus fruit detection and identification and counting. Finally, regression analysis is performed by combining known yield-related data from previous years to achieve the effect of citrus fruit yield prediction. The network consists of three parts: (1) Citrus identification and counting module built using Yolov5 based target detection algorithm~\cite{jocher2020ultralytics}, the main task of this module is to detect the photographed citrus fruit pictures, use the bounding box to frame the citrus fruits, and count the bounding box to obtain the counting results of citrus fruits;
(2) Model compression of Yolov5 by knowledge distillation algorithm to improve the model inference speed~\cite{hinton2015distilling};
(3) Citrus yield prediction module built based on linear regression model~\cite{seber2012linear}, which uses the number of citrus fruit as input through the regression model of weight and number of citrus fruit to obtain the yield prediction results of citrus fruit.

With this network, the following requirements can be met: (1) The number of citrus fruit can be directly obtained by two photos of citrus trees at different angles of maturity; (2) The number of citrus fruit obtained can achieve the effect of predicting the yield; (3) There is a good effect for the recognition of citrus fruit with small targets; (4) The overlapping fruits can be effectively recognized. (5) It can meet the effective recognition under different lighting conditions.

\section{Related Works}
\subsection{Object Detection}
The current object detection models can be broadly classified into one-stage and two-stage models \cite{redmon2016you,zhou2019objects,girshick2015fast}. The two-stage model classifies and detects the foreground and background of an image in one stage to generate RPNs \cite{he2016deep}, which improves the problem of imbalance between positive and negative samples. However, the two-stage sampling structure of the two-stage model and the process of screening RPNs lead to poor real-time performance of the two-stage model. After rapid development in recent years, the one-stage model has caught up with the two-stage model in terms of accuracy. The one-stage model directly processes the feature maps output from the backbone network, directly classifies and localizes objects, and outputs object class and location information~\cite{zhou2019objects,jocher2020ultralytics,redmon2016you}. In terms of model complexity, the single-stage model can better meet the real-time requirements of video surveillance and reduce the hardware requirements.

YOLOv1 takes the whole image as the input of the network and returns anchor points directly in the output layer to select object classes, so YOLOv1 can only detect images with the same pixels in the training and test sets. When YOLOv1's bounding box contains multiple objects, it can only detect classes with a large number of objects~\cite{redmon2016you}.YOLOv1 introduces batch normalization to address the problem of gradient explosion and gradient disappearance during recall.YOLOv2 trains the CNN network using low pixels (224 × 224) and then uses high-resolution (448 × 448) images to fine-tune the learned features, so it can adapt to multi-resolution scenes~\cite{redmon2017yolo9000}.YOLOv3 designed the dark network as a backbone network for better implementation of the classification task~\cite{redmon2018yolov3}.YOLOv3 performs multi-label cross-entropy classification for image classification to ensure that positive samples are used for the classification task. Loss regression balances the positive and negative samples. the model architectures of YOLOv4~\cite{bochkovskiy2020yolov4} and YOLOv5~\cite{redmon2016you} are very similar. The images were enhanced in several ways and the feature maps were deeply integrated.
\subsection{Knowledge Distillation}
Compressed network models can improve their performance in embedded devices and also enable faster data processing in the cloud and reduce the computational power burden on servers. The commonly accepted methods for model compression are pruning and knowledge refinement~\cite{hinton2015distilling,han2015learning}, proposed pruning connections of small weights in trained neural networks. The weights of the network obtained by sparse training can be zero to reduce the storage space~\cite{molchanov2016pruning}.Proposed to compress the model and then fine-tune the compressed model to improve the accuracy~\cite{han2015learning}. Decomposes the number of neurons in the training phase, so some neurons can be pruned to make the network model more compact~\cite{zhou2016less}.Uses L1 regularization to sparse the gamma coefficients, so the optimization objective is simpler.
Pruning techniques include minimum weight pruning based on convolutional kernel weights, information entropy pruning based on the effect of channels on loss, and channel pruning based on BN layers~\cite{liu2017learning,zhou2016less,hu2016network}.
Knowledge distillation is guided by the high accuracy of the teacher network, which can be used to guide students to learn knowledge online using small models. Distillation strategies can be divided into three types. (1) Response-based distillation. The loss of output from the teacher network is used to guide students' online learning~\cite{hinton2015distilling}. (2) Feature-based distillation. Student networks learn feature map information from the intermediate layers of the teacher network, as in Fitnets~\cite{romero2014fitnets}. (3) Relationship-based distillation: student networks learn feature information between layers of the teacher network and between samples. Distillation may lead to a decrease in accuracy when the gap between the depth and width of the teacher network and student network models is too large~\cite{mirzadeh2020improved}.Proposed an assistant teacher network to solve the problem of excessive gaps. Considering that the layer structure of YOLOv5s is smaller than that of large networks, we use a response-based distillation strategy.

\section{Proposed Method}
We build a citrus fruit quantity prediction network based on Yolov5 knowledge distillation and linear regression model. As shown in Fig.1, the network is divided into two parts, which represent the training part of the citrus fruit object detection model and the prediction part of citrus fruit yield. First, a student model is trained using knowledge distillation, and then the number of citrus fruit is evaluated by detecting citrus fruits, which is used as an input of the citrus fruit yield estimation model. Finally, regression calculation was performed on the model to obtain the estimated value of citrus fruit yield.

\begin{figure}[htbp]
\centerline{\includegraphics[width=1\linewidth]{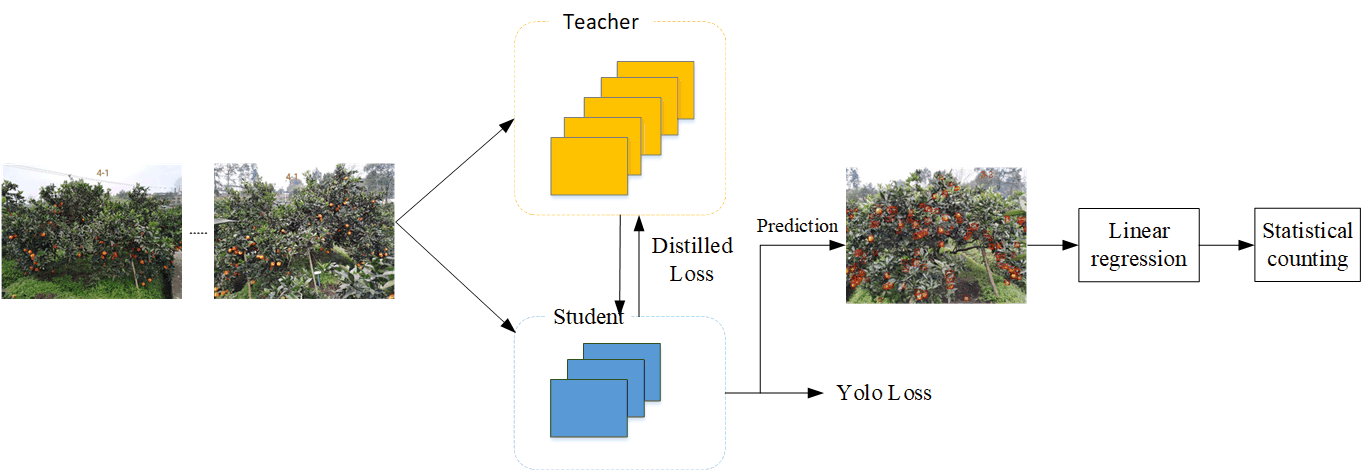}}
\caption{Citrus Yield Estimation Network}
\label{fig}
\end{figure}

\subsection{Yolov5 Detection}\label{AA}
We use Yolov5 based object detection algorithm~\cite{jocher2020ultralytics}. The network consists of Input, Backbone, Neck and Head. The input completes basic tasks such as Mosaic data enhancement, adaptive image scaling and adaptive anchor box calculation; the backbone network is used to extract image features; the neck is a network layer that combines (Concat) image features through a series of network layers, then the image features are passed to the prediction layer (the output Head); Finally, the network outputs the prediction image features, generates bounding boxes and predicts categories, and records citrus location information. 

In the Yolo detection algorithms, the Focus structure is used in the backbone network, and the key lies in the slicing operation. The 4$\times$4$\times$3 feature map is sliced into a 2$\times$2$\times$12 feature map. The 608$\times$608$\times$3 three-channel image is input into the Focus structure. After slicing operation, it becomes a 304$\times$304$\times$12 feature map, and then 32 convolution kernels are used to perform convolution operations. For different objects, the corresponding anchors with default length and width are initially set. During training, based on the initially set anchor box, a bounding box is output, and calculate the loss of the ground truth and the bounding box. 

\subsection{Knowledge Distillation}
The knowledge distillation algorithm based on Yolov5 detection is shown in Fig. 2. The algorithm is to first train a teacher network with high complexity and powerful feature extraction ability, and then it use the output of the network as soft target supervision information to guide a student network with simpler network structure and faster inference speed. To realize the transfer of the "knowledge" of the teacher network to the student network. In traditional object classification or detection tasks, data are usually marked with "0" and "1" hard labels, but the amount of information contained in hard labels is limited and does not include the relationship between classes~\cite{hinton2015distilling}. In contrast, knowledge distillation uses the output probability of the teacher network (with a value between 0 and 1) to label the data, which can well represent the similarity between different categories. 
\begin{figure}[htbp]
\centerline{\includegraphics[width=1\linewidth]{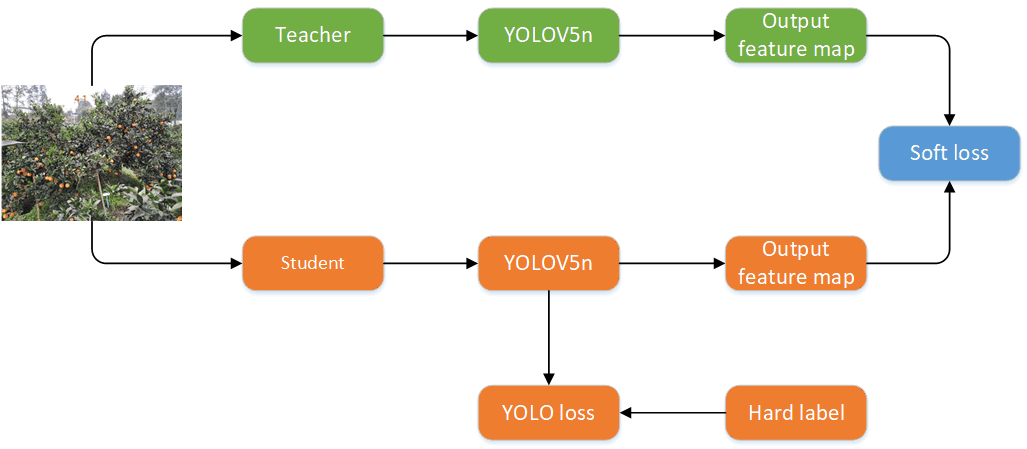}}
\caption{Yolov5 knowledge distillation training flowchart}
\label{fig}
\end{figure}

In order to obtain a "soft" classification output, the knowledge distillation algorithm introduces a temperature coefficient T into Softmax to soften the class probability of the target. The calculation formula is as follows:
\begin{equation}
q_i = exp(\frac{z_i}{T})/\sum_{j}exp(\frac{z_i}{T}) 
\end{equation}
Where $z_i$ is the content of the output layer of the neural network.
The network evaluates and predicts the network in three aspects, namely classification loss, regression loss and objectness loss, which are defined as follows:
\begin{equation}
L_{Yolo} = f_{obj}(O_{i}^{gt}, \hat{O}_i) + f_{cl}(p_{i}^{gt}, \hat{p}_i) + f_{bb}(b_{i}^{gt}, \hat{b}_i)
\end{equation}
where $\hat{O}_i$, $\hat{p}_i$, $\hat{b}_i$  are the objectness, class probability and bounding box coordinates of the student network, $O_{i}^{gt}$, $p_{i}^{gt}$ and $b_{i}^{gt}$ are the ground truth from the dataset.

\subsection{Citrus Quantity Prediction Module}
Linear regression is a commonly used mathematical analysis method, which predicts the future change or future change trend of its related random variables through the change of one or more independent variables, so as to obtain the desired prediction results. According to the number of dependent variables, linear regression can be divided into two types: univariate linear regression and multiple linear regression. In this study, the dependent variable that can be obtained at this stage is the number of citrus obtained by the target detection model. There are two dependent variables to be obtained. The first dependent variable is the single fruit weight of citrus, and the second dependent variable is the yield of citrus, where the citrus yield is the ultimate goal of this study, and the weight of single citrus fruit is the intermediate variable.

In this study, the weight of single citrus fruit will also be used as an independent variable of citrus fruit yield. The single linear regression method is used for the citrus yield prediction model, because the dependent variable and the independent variable in this study satisfy: (1) The independent variable and the dependent variable are independent of each other, and the independent variable obeys the positive (2) The variance of the dependent variable under different independent variables along the direction of the regression line is the same; (3) The relationship between the dependent variable and the independent variable is a linear relationship.

The mathematical expression of the univariate linear regression model is as follows, where $a$ and $b$ are position parameters, $b$ represents the intercept, $a$ represents the slope of the straight line, and $x^i$ is the input value. 
\begin{equation}
f(x) = ax^i+b
\end{equation}

The best curve fit by a univariate linear regression model is obtained by minimizing the error between the predicted and actual values. Mean squared error is one of the most commonly used performance measures in regression tasks, and its definition is as follows.
\begin{equation}
J(a, b)=\sum_{i=1}^n\left(f\left(x^i\right)_i-y^i\right)^2=\sum_{i=1}^n\left(a x^i+b-y^i\right)^2
\end{equation}

where $y^i$ represents the true value and $f(x^i)$ represents the predicted value. It is used in linear regression to quantify the error between the predicted result and the real result. The smaller its value, it indicates that the prediction result is closer to the target value.

To determine the values of the parameters $a$ and $b$, the method of gradient descent is used. Gradient is a vector, which means that a function changes the fastest and has the highest rate of change along a point. The gradient descent method obtains the optimal solution through multiple iterations, and its expression is as follows, where $\alpha$ is the learning rate, and this parameter controls the step size of each iteration.
\begin{equation}
\text { repeat }\left\{\theta_j:=\theta_j-\alpha \frac{\partial_J(\theta)}{\partial \theta_j}\right\}
\end{equation}

\section{EXPERIMENT RESULTS}
\subsection{Experiment Settings}
The algorithm in this paper is implemented on the open source pyTorch1.7.0 framework using Nvidia 1080Ti GPUs and the operating system is Linux. The main hyper-parameters are set as follows. We train the teacher network Y olov5s-T for 300 epochs, the batchsize is 4, the initial learning rate is set to 0.01, and the distillation temperature parameter T=20 is introduced into the output layer of the network.
\subsection{Dataset}
In order to realize the detection of citrus fruit, we selected Newhall navel orange, a representative citrus and one of the main citrus varieties in the southwest hilly area, as the research object. Image acquisition was performed with an MI 6X mobile phone. In order to simulate the variability and randomness of the application scene, the shooting height is the handheld height (about 1.5m), the mobile phone is 1.5m~2m away from the tree, and We take two full tree photos of each tree from two different angles, for a total of 40 photos of 20 Newhall trees. The photo resolution is 4000×3000 pixels, and the photo save format is jpg. The images include overlapping, occluded, small-volume citrus fruit images. The 20 trees in this experiment were selected from 5 experimental fields, and 4 trees were randomly selected from each experimental field, and were marked in the form of No.1 citrus tree in No.4 experimental field as 4-1, the image is shown in Fig.3.
\begin{figure}[htbp]
\centerline{\includegraphics[width=1\linewidth]{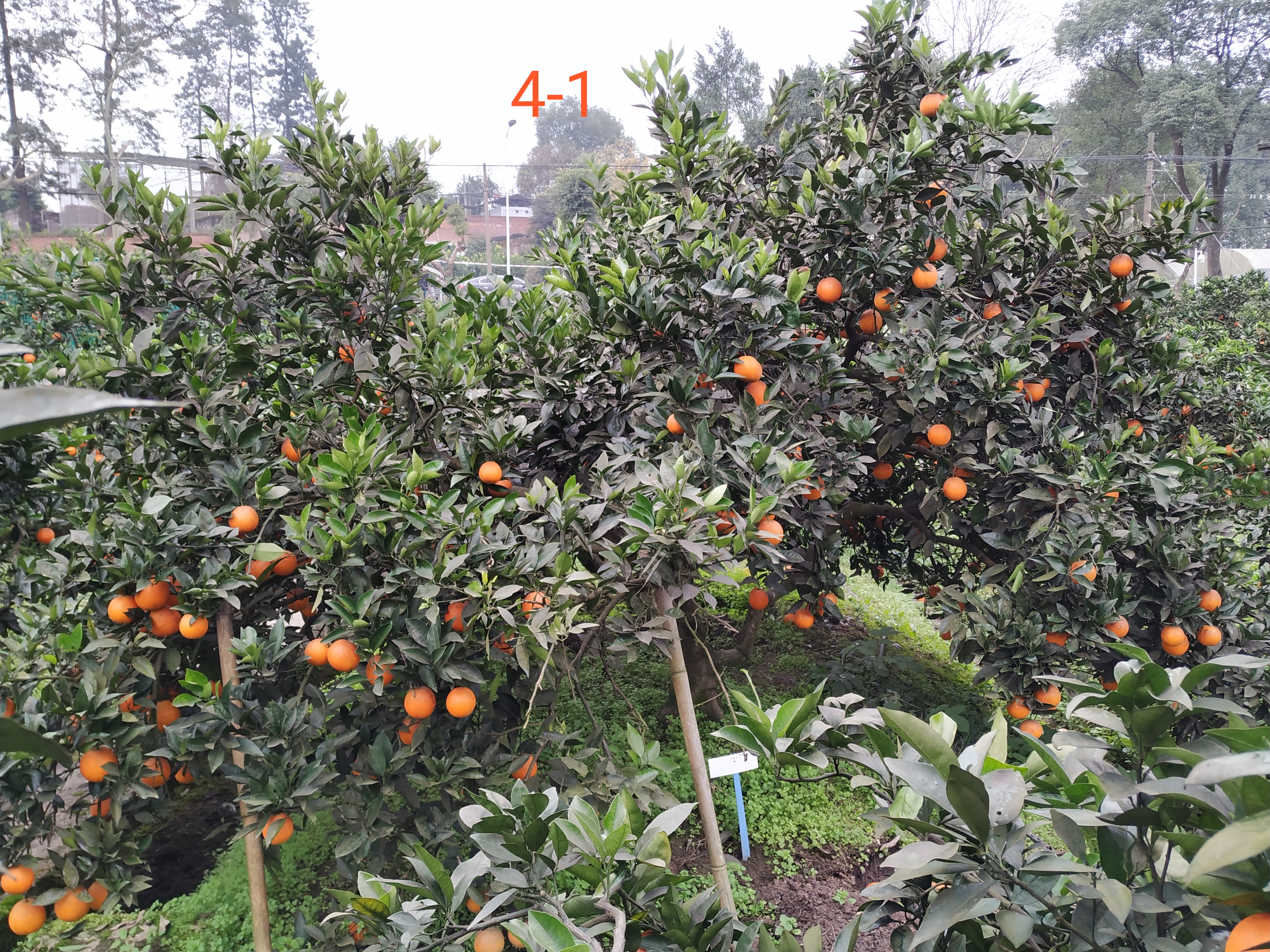}}
\caption{4-1 Citrus Tree Picture}
\label{fig}
\end{figure}
\subsection{ Experiment Analysis and Ablation Study}
In order to verify the performance of the above proposed Yolov5 algorithm based on knowledge distillation, ablation experiments are carried out. In this paper, the designed student network is named Yolov5n-S, and the teacher network is named Yolov5s-T. The basic parameter information of the student network is the same as that of the teacher network. The test results are shown in the Tab.1.
\begin{table}[htbp]
\caption{Analysis on knowledge distillation }
\begin{center}
\begin{tabular}{|c|c|c|c|c|c|}
\hline
\textbf{Net}&{\textbf{P/\%}}&{\textbf{R/\%}}&{\textbf{MAP/\%}} & {\textbf{P$_{arams}$/M}}\\
\hline
\textbf{Yolov5s-T} & {84.9}& {74.8}& {83.3} & {15.11} \\
\hline
\textbf{Yolov5s-S} & {77.8}& {88.4}& {83.6} & {13.43} \\
\hline
\end{tabular}
\label{tab1}
\end{center}
\end{table}

It can be seen from the test results that after the introduction of the knowledge distillation algorithm, the size of the model (P$_{arams}$) is reduced from the original 15.11MB to 13.43MB, which is significantly reduced to 88.9$\%$ of the original, and the recall rate (Recall) is increased from 74.8$\%$ of the teacher model to 13.43MB. 88.4$\% $improves by 13.6$\%$, MAP value remains almost unchanged, and the precision (P) decreases from 84.9$\%$ for teacher network to 77.8$\%$. Accuracy drops slightly.

It can be seen that the actual number of identifications of each tree is not much different from the number of markers, the average relative error is 18$\%$, the largest relative error of a single fruit tree is 13$\%$, and the smallest relative error is 0$\%$. Among them, the total number of labeled fruit instances is 5069, and the total actual recognition number is 4675. The accuracy rate is 86.56$\%$. The IOU used in this experiment is 0.5, that is, the threshold of the intersection ratio of the predicted box and the ground truth is 50$\%$, the confidence level used is 0.15, and AP, Recall and Precious evaluations are performed on the test images of each tree. The average precision of all samples is about 78$\%$, the average AP is about 84$\%$, and the average recall is about 88$\%$, and the detection effectiveness is ideal, as shown in Fig.4. One of the evaluation results is shown in Fig. 4.

\begin{figure}[htbp]
\centerline{\includegraphics[width=1\linewidth]{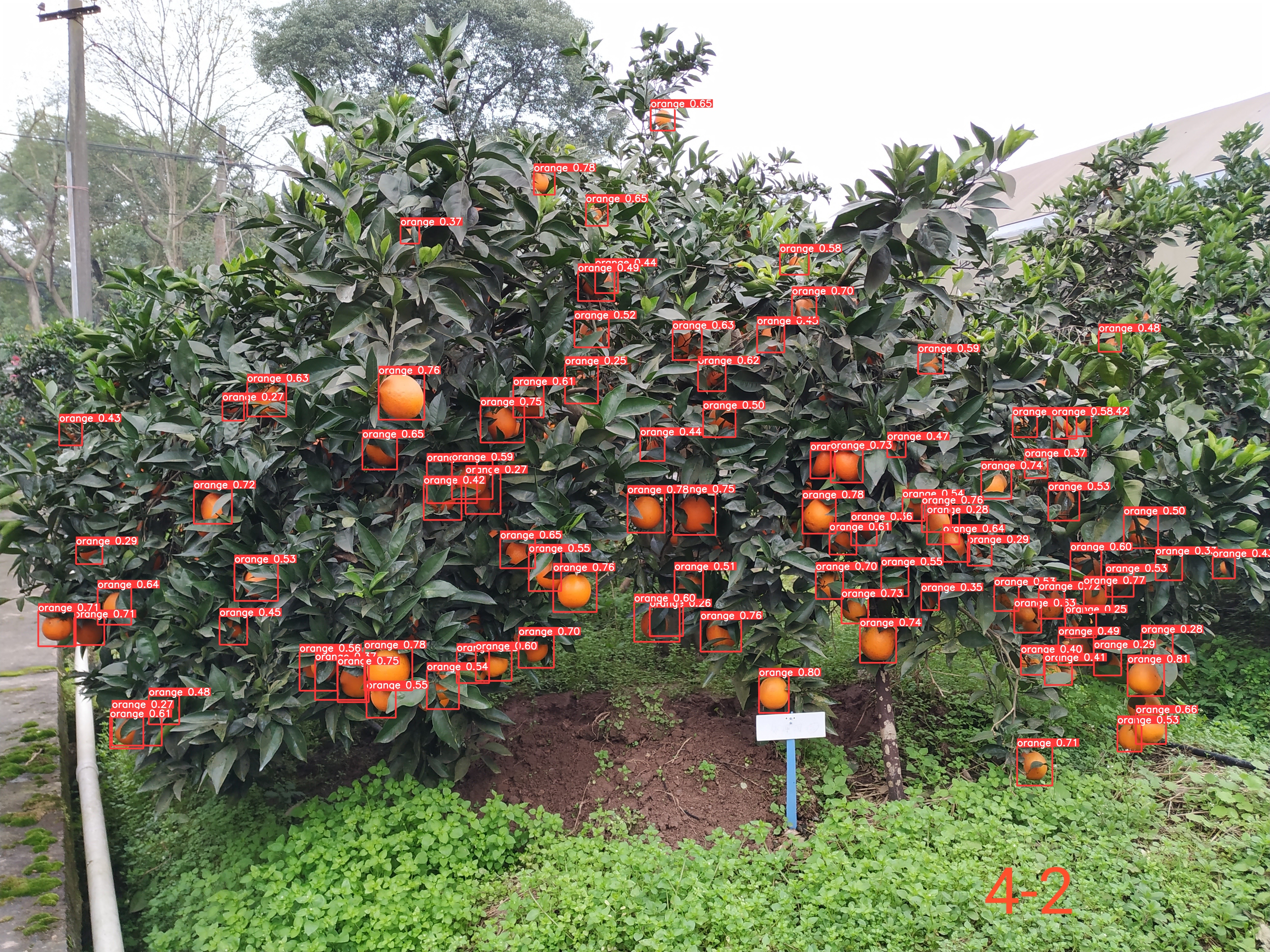}}
\caption{Visualization of inspection results}
\label{fig}
\end{figure}

\begin{figure}[htbp]
\centerline{\includegraphics[width=1\linewidth]{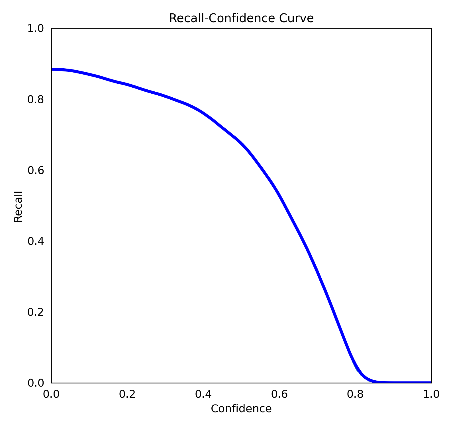}}
\caption{Validation set of Recall}
\label{fig}
\end{figure}

\begin{figure}[htbp]
\centerline{\includegraphics[width=1\linewidth]{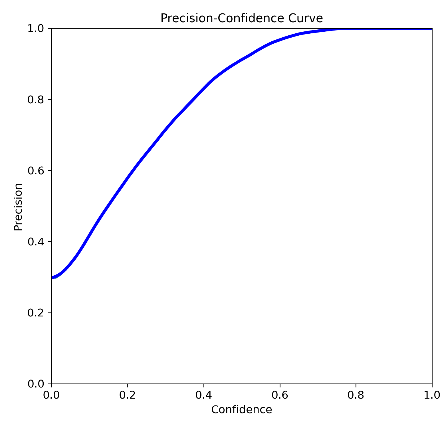}}
\caption{Validation set of Precision}
\label{fig}
\end{figure}

To analyze the above figure, we also can see that the Recall and Precision of the detection results are negatively correlated, with up and down fluctuations in local areas. As can be seen from the AP curve, the curve is closer to the (1, 1) point in the upper right corner, indicating that our model has a good recall and accuracy rate overall.
Based on the model building part, the number of citrus fruits detected is very close to the actual number of marked citrus fruits due to the ideal success rate of the Yolov5-knowledge distillation target detection algorithm for detecting citrus fruits. In order to build the model of citrus number, this study fitted the actual number of tagged to the real citrus fruits single tree citrus number by doing a univariate linear regression and obtained the citrus fruits number model: y = 0.998 x - 15.101, where x is the actual number of tagged to citrus and y is the real single tree number of citrus.

\begin{table}[!h]
    \caption{Identification of the number of citrus fruits for each sample after linear fitting}
  
    \centering
    \begin{tabular}{cccc}
        \hline
 No. &Estimated number & Ground Truth & Relative error of detection\\
        \hline
  1	&75	&53	&41.5$\%$ \\
2	&74	&40 &	85.0$\%$  \\
3	&73&	50	&46.0$\%$ \\
4	&90&	 67&	34.3$\%$ \\
5	&98  &	81&	21.0$\%$ \\
6	&85	  &70	&21.4$\%$ \\
7	&126	&108	&16.7$\%$ \\
8	&180	&156	&15.4$\%$ \\
9	&135	&125	&8.0$\%$\\
10	&151	&141	&7.0$\%$\\
11	&118	&97&21.6$\%$\\
12	&103	&88	&17.0$\%$\\
13	&104	&85	&22.3$\%$\\
14	&105	&88	&19.3$\%$\\
15	&94  &	98	&4.1$\%$\\
16	&112	&108	&3.7$\%$\\
17	&141	&135	&4.4$\%$\\
18	&132	&132	&0.0$\%$\\
19	&154	&133 &	15.7$\%$\\
20	&130	&115	&13.0$\%$\\

        \hline       
    \end{tabular}
    \label{bs2}
\end{table}

We used the relative error formula to evaluate the prediction results of the yield. The relative error formula is as follows, where P represents the predicted value and T represents the true value.

$$E=\frac{\lvert{P-T}\rvert}{T}$$

As can be seen from the above table, the absolute error between the estimated number of individual citrus trees and the true yield of individual citrus trees for the second citrus tree varies between 0\% and 30\%, with significant fluctuations. However, the relative error for the overall yield of 20 citrus trees was 8.5\%. The above comparison shows that the use of a linear regression model is more effective in predicting the total yield of citrus. The data set used in this study is small, and if the data set is increased appropriately, the yield prediction effectiveness for single citrus tree should be improved accordingly.

\section{CONCLUSION}
We use the Yolov5 detection algorithm for the location of citrus fruits. At the same time, in order to improve the inference speed of the model and reduce the amount of model parameters, we use the knowledge distillation method, introduce the student network to compress the model, and obtain the predicted number of citrus fruits and citrus fruits according to statistics. There is a linear relationship between the actual numbers, and a linear regression model is used for the counts and yield estimates of citrus fruit trees.


\end{document}